# Predicting 1p19q Chromosomal Deletion of Low-Grade Gliomas from MR Images using Deep Learning

Zeynettin Akkus, Issa Ali, Jiří Sedlář, Timothy L. Kline, Jay P. Agrawal, Ian F. Parney, Caterina Giannini, Bradley J. Erickson

*Abstract*— **Objective:** Several studies have associated codeletion of chromosome arms 1p/19q in low-grade gliomas (LGG) with positive response to treatment and longer progression free survival. Therefore, predicting 1p/19q status is crucial for effective treatment planning of LGG. In this study, we predict the 1p/19q status from MR images using convolutional neural networks (CNN), which could be a noninvasive alternative to surgical biopsy and histopathological analysis. **Method:** Our method consists of three main steps: image registration, tumor segmentation, and classification of 1p/19q status using CNN. We included a total of 159 LGG with 3 image slices each who had biopsy-proven 1p/19q status (57 nondeleted and 102 codeleted) and preoperative postcontrast-T1 (T1C) and T2 images. We divided our data into training, validation, and test sets. The training data was balanced for equal class probability and then augmented with iterations of random translational shift, rotation, and horizontal and vertical flips to increase the size of the training set. We shuffled and augmented the training data to counter overfitting in each epoch. Finally, we evaluated several configurations of a multi-scale CNN architecture until training and validation accuracies became consistent. **Results:** The results of the best performing configuration on the unseen test set were 93.3% (sensitivity), 82.22% (specificity), and 87.7% (accuracy). **Conclusion:** Multi-scale CNN with their self-learning capability provides promising results for predicting 1p/19q status noninvasively based on T1C and T2 images. **Significance:** Predicting 1p/19q status noninvasively from MR images would allow selecting effective treatment strategies for LGG patients without the need for surgical biopsy.

## I. Introduction

Magnetic resonance (MR) imaging is a non-invasive medical imaging technique that provides outstanding soft tissue contrast and has become the standard imaging technique for brain tumor diagnosis [1]. Gliomas are the most frequent primary brain tumors originating in the brain [2]. World Health Organization (WHO) classifies them into four grades based on their aggressiveness. Low-grade gliomas (LGG), also named diffuse low-grade and intermediate-grade gliomas (WHO grades II and III), include oligodendrogliomas, astrocytomas, and oligoastrocytomas [3-5]. Compared to high-grade gliomas (HGG: WHO grade IV, glioblastoma), LGG are less aggressive tumors with better prognosis. A subgroup of LGG will progress to glioblastoma (HGG, grade IV), but other subgroups will progress slower or remain stable [4, 6-8]. In addition, some LGG are sensitive to therapy and their survival ranges from 1 to 15 years [6, 8]. Presently, treatment includes observation, surgery, radiotherapy, and chemotherapy either separately or in combination [4]. Although histopathological study is the gold standard for diagnosis and subsequent treatment planning, it is known that histopathological diagnosis lacks information about other tumor properties that can impact optimal therapy options. Therefore, other tests of LGG (e.g. molecular biomarkers testing) are also obtained to improve treatment planning. Several studies [9-12] have shown that codeletion of 1p/19q chromosome arms is a strong prognostic molecular marker for positive tumor response to chemotherapy and radiotherapy in LGG and associated with longer survival. Therefore, predicting 1p/19q status is crucial for effective treatment planning of LGG.

Currently, determining 1p/19q status requires surgical biopsy typically followed by fluorescence in-situ hybridization (FISH) [13] to identify chromosomal deletion. Several studies have shown that imaging can predict 1p19q status from MR images or positron emission tomography (PET) images. Fellah et al. [9] presented univariate analysis and multivariate random forest models to determine 1p/19q status from multimodal MR images including conventional MR images, diffusion weighted imaging (DWI), perfusion weighted imaging (PWI), and MR spectroscopy. DWI, PWI, and MRI spectroscopy showed no significant difference between tumors with and without 1p/19q loss in their study. They concluded that inclusion of DWI, PWI, and MR spectroscopy was not useful for determining 1p/19q status compared with

This work was supported by National Institutes of Health 1U01CA160045. TLK is supported by the National Institute of Diabetes and Digestive and Kidney Diseases (NIDDK) under Grant/Award Number P30 DK090728, and the PKD Foundation under grant 206g16a.
Z. Akkus, I. Ali, J. Sedlar, T.L. Kline, J. Agrawal, and B. J. Erickson are with Radiology Informatics Lab, Mayo Clinic, Rochester, MN, USA (e-mail: zetgate@gmail.com, ali.issa@mayo.edu, sedlar.jiri@mayo.edu, kline.timothy@mayo.edu, doctor.jpagrawal@gmail.com, bje@mayo.edu ). Caterina Giannini is with Department of Pathology, Mayo Clinic, Rochester, MN, USA (e-mail:Giannini.Caterina@mayo.edu). Ian F. Parney is with Department of Neurologic Surgery, Mayo Clinic, Rochester, MN, USA (e-mail: Parney.Ian@mayo.edu)

conventional MR images. Jansen et al. [10] presented detection of 1p/19q status from [$^{18}$F] fluoroethyltyrosine-PET (FET-PET) images. They derived several biomarkers from PET images and correlated with 1p/19q status, and showed that these biomarkers do not reliably predict the status of 1p/19q in individual patients. Iwadate et al. [11] studied detection of 1p/19q codeletion from $^{11}$C-methionine PET images and concluded that $^{11}$C-methionine PET might help discriminate tumors with and without 1p/19q codeletion preoperatively. Bourdillion et al. [12] presented prediction of anaplastic transformation in grade 2 oligodendrogliomas based on MR spectroscopy and 1p/19q status. They showed that choline/creatine ratio >2.4 was associated with the occurrence of anaplastic transformation in patients without 1p/19q codeletion. On the other hand, no anaplastic transformation was observed in patients with 1p/19q codeletion.

In this study, we present a robust, and noninvasive method to predict the 1p/19q status of LGG from post-contrast T1 and T2 weighted MR images using convolutional neural networks (CNN).

## II. MATERIAL AND METHODS

We use a combination of two commonly acquired image types, T2 and post-contrast T1 weighted images, as input to our classification algorithm. Figure 1 shows examples of LGG image characteristics in post-contrast T1 and T2 weighted images with and without 1p/19q codeletion. Our classification algorithm consists of several pre-processing steps: Multi-modal image registration, tumor segmentation, data normalization, and data augmentation. After applying pre-processing steps to data, we use the segmented images to train a multi-scale CNN for prediction of 1p/19q status. A flowchart and implementation details of the multi-scale CNN are shown in Figure 2.

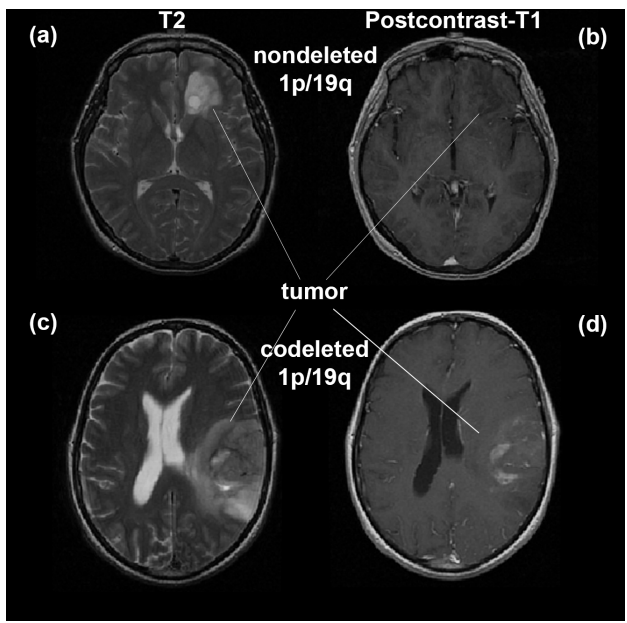

Figure 1: An example of low-grade glioma with and without 1p/19q codeletion. Images a and b show T2 and post contrast T1 for nondeleted 1p/19q. Images c and d show T2 and post contrast T1 for codeleted 1p/19q.

### A. Brain Tumor Data

One hundred fifty nine (n=159) consecutive (01-Oct-2002 – 01-Aug-2011) pre-operative LGG patients, with stereotactic MRI images, who had biopsy proven 1p/19q status consisting either no deletion or co-deletion, were identified from our brain tumor patient database at Mayo Clinic for this study. The data included 102 nondeleted and 57 codeleted LGG. The types of LGG were oligoastrocytoma (n = 97), oligodendrogliomas (n = 45), and astrocytomas (n = 17). In total, 477 slices (3 slices per LGG including one centered at the tumor 'equator' plus one slice above and below) were included for this study. Post-contrast T1 and T2 weighted images were available for all selected patients. Institutional Review Board (IRB) approval was obtained for this study and requirement for patient consent was waived.

All images were acquired for biopsy planning purposes, and so a very consistent scanning protocol was used, including 3mm thick T2 images and 1mm thick axial spoiled-gradient recalled images all acquired at 1.5T or 3T on either General Electric Medical System (Waukeshaw, WI) or Siemens Medical System (Malvern, PA) scanner.

### B. Pre-processing

*Multi-modal Image Registration*

We registered post-contrast T1 weighted images to T2 weighted images of the same patient by using the ANTs open source software library for image registration [14, 15]. We performed rigid registration using cubic b-spline interpolation and mutual information metric, which takes into account translational movements, to align our intra-patient images.

*Tumor Segmentation*

We used our semi-automatic LGG segmentation software to segment the tumors in 2D [16]. First, the user selects the slice where the area of the tumor appears largest, and then draws a region-of-interest (ROI) that completely encloses the tumor and some normal tissue. Second, a normal brain atlas [17] and post-contrast T1 weighted images are registered to T2 weighted images. Third, the posterior probability of each pixel/voxel belonging to normal and abnormal tissues is calculated based on information derived from the atlas and ROI. Finally, geodesic active contours [18] use the probability map of the tumor to shrink the ROI until optimal tumor boundaries are found. With that, a morphological binary dilation of five pixels is applied to make sure boundaries are included in the tumor ROI.

*Data Normalization*

After registration and segmentation steps, the dataset of raw MR image data is normalized to balance intensity values and narrow the region of interest. This step aims to finely tune the input information fed into the CNN. The normalization process begins with skull-stripping using Brain Extraction Tool (BET) [19] based on FSL library [20], then followed by standard scoring. Standard scores (also called z-scores) are calculated for each image by subtracting the mean of image intensities from an individual intensities and then dividing the difference by the standard deviation of the image intensities (see Eq. 1).

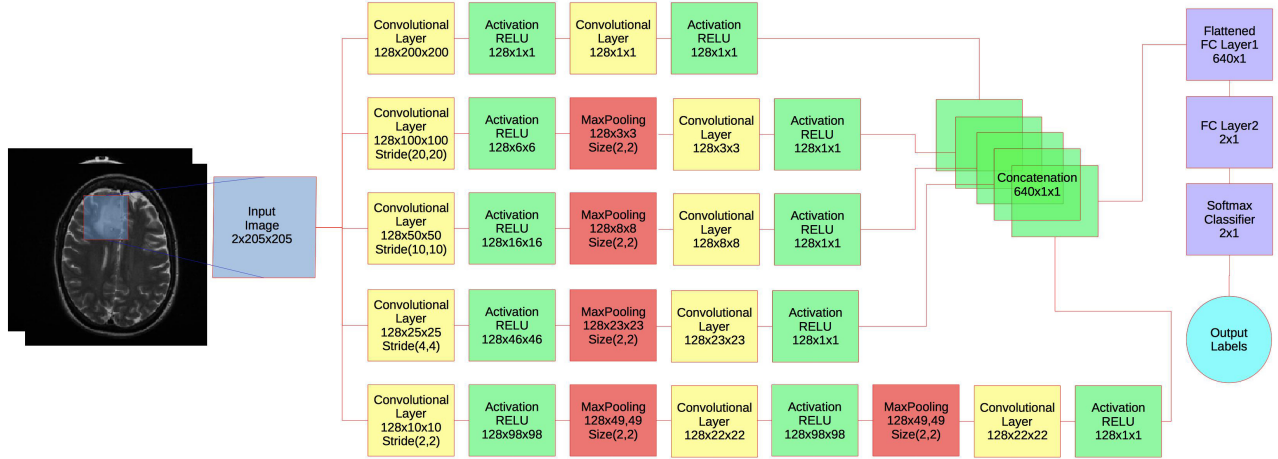

Figure 2: A flowchart of the multi-scale CNN architecture. Blue box is the input image. Yellow boxes are convolutional layers. Green boxes are rectified linear units (RELU), activations. Red Boxes are max pooling layers. Purple boxes are fully connected layers plus a softmax binary classifier. Cyan circle shows the output labels.

$$z = \frac{X - \mu}{\sigma}$$

Eq.1

where $X$ is image intensities, $\mu$ is mean of the image intensities, and $\sigma$ is the standard deviation of the image intensities.

*Data Augmentation*

Using deep networks, in particular CNNs, there is a high susceptibility to overfitting. This is a direct result of the large number of network parameters relative to the number of features provided by the MR images. The number of features available from MR images may not be sufficient to provide sufficient learning and generalizability to the parameters in the network, hence the need to increase the number of MR images. One approach to address this is through artificially augmenting the dataset using label-preserving transformations [21]. This "data augmentation" consists of generating image translations, rotations, and horizontal and vertical flipping. We apply a random combination of these transformations on each image, hence creating "new" images. This multiplies our dataset by several fold and help reduce over-fitting.

*C. Convolutional Neural Networks*

Convolutional Neural Networks (CNN) is a type of feed-forward artificial neural network for learning a hierarchical representation of image data [22]. Unlike a regular Neural Network, the layers of a CNN have neurons arranged in 3 dimensions (width, height, and depth) and respond to a small region of the input image, called receptive field, instead of all of the neurons in a fully-connected manner. Each neuron learns to detect features from a local region the input image. This allows capturing features of local structures and preserving the topology of the input image. The final output layer will reduce the full image into a single vector of class scores, arranged along the depth dimension.

The main types of layers needed to build a CNN deep learning system are: input layer, convolutional layer, activation layer, pooling layer, and fully connected layer. Most implementations have many of each type of layer, hence the title 'deep'learning. Each layer is described in more detail.

Input layer: This layer holds the raw pixels values of the input image after applied pre-processing steps.

Convolutional layer: This layer is composed of several feature maps along the depth dimension, each corresponding to a different convolution filter. All neurons with the same spatial dimension (width and height) are connected to the same receptive filed in the input image or, generally, in the previous layer. This allows capturing a wide variety of imaging features. The depth of the layer, i.e. the number of convolution filters, defines the number of features that can be extracted from each input receptive field. Each neuron in a feature map shares exactly the same weights, which define the convolution filter. This allows reducing the number of weights, and thus increasing the generalization ability of the architecture.

Activation layer: This layer applies an activation function to each neuron in the output of the previous layer. For example, rectified linear unit (RELU) where RELU(x)= max(0,x) is the most common activation function used in CNNs architectures and fires the real value of the output and thresholds at zero. This layer does not change the size of the previous layer. It simply replaces negative values with '0'.

Pooling layer: Placed after an activation layer and this layer down-samples along spatial dimensions (width and height). It selects the invariant imaging features by reducing the spatial dimension of the convolution layer. The most popular type is max pooling, which selects the maximum value of its inputs as the output, thus preserving the most prominent filter responses.

Fully connected layer: As with neural networks, this layer connects all neurons in the previous layer to this layer with a weight for each such connection. If used as the output, each output nodes represents the 'score' for each possible class.

To allow learning of complex relationships and to achieve a more hierarchical representation of the input image, multiple convolutional-pooling layers are stacked to form a deep architecture of multiple nonlinear transformations. This allows learning a hierarchy of complex features carrying predictive power for image classification tasks.

*Multi-Scale CNN parameters*

In our study, we use a CNN architecture that consists of different sizes of convolutional filters to train on our dataset (see Figure 2). This can be considered as multi-scale CNN that learns global and local imaging features with different convolutional filter sizes and concatenates their output before the classification step. The parameters of each layer are seen in the boxes in Figure 2. For example, the size of the first convolutional layer in the first branch is 128x200x200 and corresponds to depth of layer, 128 filters, and size of filters 200 by 200 in 2D spatial space (**x**, **y**). Default stride size is 1 pixel in **x** and **y**. Specific strides for each layer are shown in the boxes. Since every tumor has a different shape and size, we embedded each 2D tumor slice into standard background (zero intensity level) of size 205 by 205 pixels, which is the smallest that encompasses the largest tumor size in our data set.

In our architecture, probability of tumor slices belonging to each class were computed with softmax classifier and the parameters of the CNN were updated by minimizing a negative log likelihood loss function [23]. We used a stochastic gradient descend (SGD) [24] algorithm with mini batches of 32 samples in our study. The SGD algorithm with mini batches is commonly used to train neural networks on large datasets because it efficiently finds good values without high computation or memory requirements. Specifically, only a small batch of training data at a time is used at each update of weights instead of using all training samples to compute the gradient of the loss function. The learning rate was initially set to 0.001 and decreased 50% at every 50 epoch. The training was stopped when the change in validation loss smaller 0.02 for 10 consecutive epochs.

III. IMPLEMENTATION AND EXPERIMENTS

*Implementation*

Our code is based on the Keras package [25], built on top of Theano library, a Python library. Keras can leverage Graphical Processing Units (GPUs) to accelerate the deep learning algorithms. We trained our CNN architecture on an NVIDIA GTX 970 GPU card. The training took about 10 minutes to 2 days depending on the amount of the used data. For example, using 30-fold augmented data as the training data at each epoch took about 2 days to train the network.

*Experiments*

We divided our data (n=477 slices) into training and test sets. A total of 90 slices (45 nondeleted and 45 codeleted) were randomly selected from the data at the beginning, as a test set, and never seen by the CNN during the training. From the remaining data (n=387), 252 slices that were balanced for equal class probability (n = 126 codeleted + 126 nondeleted) were randomly selected at each epoch for training. Twenty percent of the training set was separated as validation set (n = 68) during the training. The data augmentation was applied to the training set (n = 7560 for 30-fold augmentation) at each epoch to increase the training samples and achieve generalization ability.

We used the performance of our CNN architecture on validation set to tune the hyper-parameters of the CNN. Since our architecture includes multi-size of convolutional filters and multi-branch CNN, it takes into account a range of values for hyperparameters in the architecture. For further tuning of hyperparameters, we first investigated the contribution of each image channel for prediction of 1p/19q status. Therefore we experimented three configurations of our multi-scale CNN based on combinations of input images without data augmentation, i.e. T1 only (Configuration 1), T2 only (Configuration 2), and T1 and T2 combined (Configuration 3). We selected the best configuration (Configuration 3) among these based on their performance on the test set. Next, we trained the best configuration from above with multiple folds $k \in \{10,20,30\}$ of data augmentation to tune the $k$ hyper-parameter of the data augmentation. We also compared the performances of our CNN architecture using four different optimizers: SGD, root mean square propagation (RMSprop), improved adaptive gradient algorithm (AdaDelta), and adaptive moment estimation (Adam). Configuration 4 was defined as the one using the best performing parameters. For each configuration, sensitivity, specificity, and accuracy were computed as follows:

Sensitivity or True Positive Rate ($TPR$):
$$TPR = \frac{TP}{TP + FN}$$
where TP is true positives, FN is false negatives

Specificity ($SPC$) or True Negative Rate:

$$SPC = \frac{TN}{TN + FP}$$
where TN is true negatives, FP is false positives

and accuracy ($ACC$):

$$ACC = \frac{TP + TN}{TP + FP + TN + FN}$$

## IV. RESULTS

The results (Figure 3) show that the multi-scale CNN is overfitting to the original (limited size) data, when data augmentation is not used, The accuracy approached 100% after 50 epochs for both the training and validation sets as seen in Figure 3 and Table 1 and 2, but remains below 80% for the test data (see Table 3). As seen in the Figure 4, the loss also stops decreasing after 150 epochs for both the training and validation sets. As seen in the Table 3, the performances of the CNN configurations without data augmentation are lower than the other configurations.

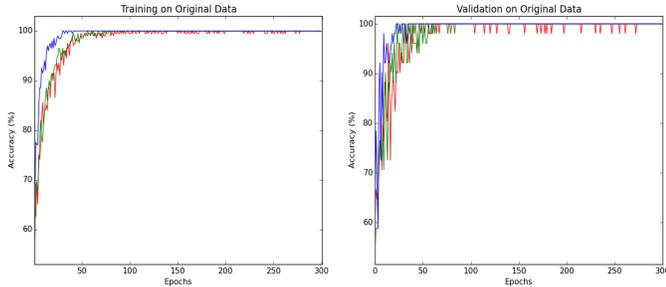

Figure 3: Accuracy plots are shown for the training (left) and validation (right) sets on the original data. Blue line corresponds to the use of T1C and T2 combined as the input image (configuration 3). Green line corresponds to the use of T2 only as the input image (configuration 2). Red line corresponds to the use of T1C only as the input image (configuration 1).

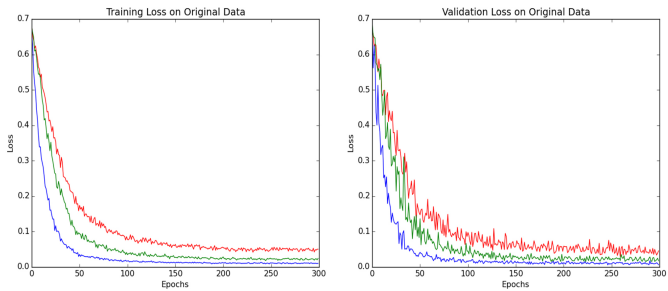

Figure 4: Loss plots are shown for the training (left) and validation (right) sets on the original data. Blue line corresponds to the use of T1C and T2 combined as the input image (configuration 3). Green line corresponds to the use of T2 only as the input image (configuration 2). Red line corresponds to the use of T1C only as the input image (configuration 1).

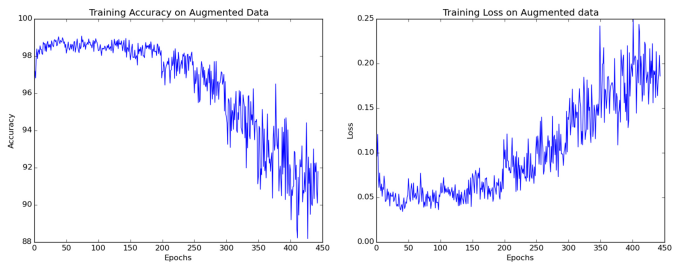

Figure 5: Accuracy (left) and loss (right) plots are shown for the training of the best performing configuration (4) on the augmented data.

Data augmentation of original data with $k$=30-fold ($ACC = 89.5$) gave the better accuracy on the validation dataset compared to the ones using $k$=10-fold ($ACC = 85.5$) and $k$=20-fold ($ACC = 83.3$) augmentations. Compared to training on the data without augmentation as seen in Figure 3 and 4, accuracy and loss in Figure 5 fluctuate and are noisier. As seen in Table 3, the performance of the CNN configuration 4 using the data augmentation are higher than the ones using the original data.

The Configuration 4 using T1C and T2 combined as the input image with the 30-fold data augmentation and SGD optimizer gives the best performance on the test data. Figure 5 shows the performance of the best performing CNN configuration 4 on the training data. Table 1, 2 and 3 shows the performances of the CNN configurations on the training, validation and test datasets. Table 4 shows the performance of the best performing configuration using 4 different optimizers. Figure 6 shows the validation loss of the best performing configuration 4 on the validation set.

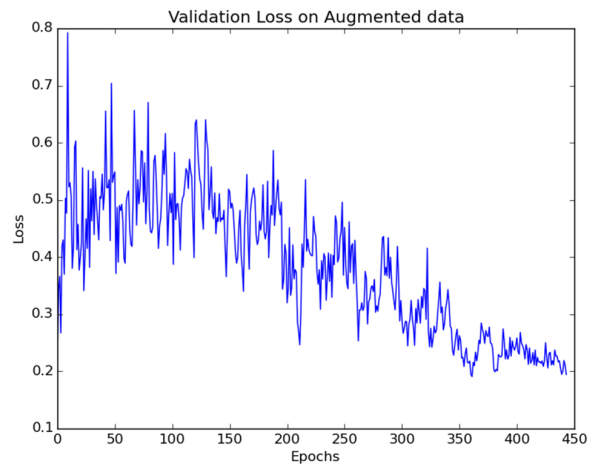

Figure 6: Validation loss is shown on the validation set for the best performing configuration (configuration 4).

| Table 1: Statistics for Training Set | | | |
|---|---|---|---|
| Configurations | Sensitivity | Specificity | Accuracy |
| 1 | 100.0% | 100.0% | 100.0% |
| 2 | 100.0% | 100.0% | 100.0% |
| 3 | 100.0% | 100.0% | 100.0% |
| 4 | 94.5% | 92.7% | 91.7% |

Table 1: The sensitivity, specificity, and accuracy for each configuration of multi-scale CNN for the training set. Configurations: 1- using T1C only and no augmentation (NA), 2-using T2 only and NA, 3: using T1C and T2 combined (T1T2) and NA, 4-using T1T2 and 30-fold augmentation.

| Table 2: Statistics for Validation Set | | | |
|---|---|---|---|
| Configurations | Sensitivity | Specificity | Accuracy |
| 1 | 100.0% | 100.0% | 100.0% |
| 2 | 100.0% | 100.0% | 100.0% |
| 3 | 100.0% | 100.0% | 100.0% |
| 4 | 89.7% | 89.4% | 89.5% |

Table 2: The sensitivity, specificity, and accuracy for each configuration of multi-scale CNN for the validation set. Configurations: 1- using T1C only and no augmentation (NA), 2-using T2 only and NA, 3: using T1C and T2 combined (T1T2) and NA, 4-using T1T2 and 30-fold augmentation.

| Table 3: Statistics for Test Set | | | |
|---|---|---|---|
| Configurations | Sensitivity | Specificity | Accuracy |
| 1 | 80.0% | 46.7% | 63.3% |
| 2 | 86.7% | 64.4% | 75.6% |
| 3 | 84.4% | 73.3% | 78.9% |
| 4 | 93.3% | 82.2% | 87.7% |

Table 3: Table shows sensitivity, specificity, and accuracy for each configuration of multi-scale CNN for the test set. Configurations: 1- using T1C only and no augmentation (NA), 2-using T2 only and NA, 3: using T1C and T2 combined (T1T2) and NA, 4-using T1T2 and 30-fold AG and further training.

| Table 4: Statistics for Optimizers | | | |
|---|---|---|---|
| Configurations | Sensitivity | Specificity | Accuracy |
| SGD | 93.3% | 82.2% | 87.7% |
| RMSprop | 84.4% | 84.4% | 84.4% |
| AdaDelta | 82.2% | 84.4% | 83.3% |
| Adam | 88.8% | 82.2% | 85.5% |

Table 4: The performance of Configuration 4 with four different optimizers on the test dataset was shown.

## V. DISCUSSION

In this study, we present a robust and noninvasive method to predict 1p/19q chromosomal arms deletion from post-contrast T1 and T2 weighted MR images using multi-scale CNN approach. As mentioned in previous studies [4, 6-8], a subset of LGG are sensitive to therapy and have longer progress free survival while others may progress to HGG. Although histologic grade is the most important factor in therapeutic decision-making, it lacks information about other tumor properties that can impact optimal therapy options and recent reports suggest genomic marker may be more predictive than histologic grade [26]. Adding other information such as 1p/19 chromosomal arms deletion, which has been associated with positive response to therapy, could help improve therapeutic decision-making.

An important challenge in applying deep learning methods to medical images is having an adequate number of data sets. Our multi-scale CNN had 100% sensitivity, specificity, and accuracy in training and validation sets without data augmentation (see Table 1 and 2: Configurations 1-3 and Figure 3), but had less than 80% specificity and accuracy for the (unseen) test set (see Table 3). This demonstrated that our algorithm did not truly learn the distinctive imaging features, but rather, overfit to the training examples. As seen in Table 3, configuration 4 using augmented data to train our multi-scale CNN approach perform better than the other configurations on the unseen test dataset. Fluctuations as seen in Figure 5 are results of introducing new augmented training data at each epoch. As seen in the results, augmenting a randomly selected set of the original training data, which is described in the experiment section, at each epoch improves generalization ability of the multi-scale CNN and prevents overfitting. As seen in Table 3, configuration 4 yields the best performance on the test dataset. The performance of configuration 4 in the training and validation datasets is in the same order with its performance in the test dataset. This highlights the generalizability of configuration 4 and its minimal overfitting.

The misclassification rate on the test dataset is about 11%. Some part of this error might be due to the error rate introduced by the FISH test that was used to determine the 1p/19 status. Scheie et al. [13] showed that the reliability of FISH test was 95% and 87.5% for the detection of 1p and 19q deletions, respectively.

As seen in Figure 6, the loss on validation dataset for configuration 4 stops fluctuating and stabilizes with slight changes after 400 epochs for several epochs. This means that the network not learning much any more and had converged to a steady state. As seen Table 4, SGD optimizer gave superior performance on the test dataset compared to other three optimizers.

As mentioned before, Fellah et al. [9] presented the first study that determined the 1p/19q status from DWI, PWI, MR spectroscopy and conventional MR images. Their misclassification rates were 48% and 40% for using only the conventional MR images and using all multimodal images, respectively. Compared to their results, our best performing configuration (4) was superior, with 93.3% sensitivity, 82.2% specificity, and 87.7% accuracy in the unseen test dataset. Moreover, our data was evaluated in a larger dataset (159 LGG vs. 50 LGG). To the best of our knowledge, our study is the first using the deep learning to predict 1p19q from MR images in LGGs.

Although the original data size was limited, artificially augmenting data helped increase the volume of our data. It may be that further performance gains will be realized with larger patient populations with more heterogeneous data. Furthermore, including weight regularization such as L1 or L2 in the network might improve generalizability of our networks. Further studies with larger patient populations are required to investigate these and confirm our current findings.

## VI. CONCLUSIONS

Our multi-scale CNN approach provides promising results for predicting 1p/19q codeletion status noninvasively, based on post-contrast T1 and T2 images. Our presented method could be further improved and potentially used as an alternative to surgical biopsy and pathological analysis for predicting 1p/19q codeletion status.